\documentclass[a4paper,fleqn]{cas-dc}

\usepackage[numbers]{natbib}
\usepackage{graphicx}
\usepackage{amsmath,amssymb,amsfonts}
\usepackage{enumerate}
\usepackage{booktabs}  
\usepackage{threeparttable}
\usepackage{multirow}
\usepackage{graphicx}
\usepackage{textcomp}
\usepackage{xcolor}
\usepackage{algorithm}
\usepackage{algpseudocode}
\usepackage{float}
\usepackage{lipsum}

\def\tsc#1{\csdef{#1}{\textsc{\lowercase{#1}}\xspace}}
\tsc{WGM}
\tsc{QE}
\tsc{EP}
\tsc{PMS}
\tsc{BEC}
\tsc{DE}

\begin{document}
\begin{sloppypar}
\let\WriteBookmarks\relax
\def\floatpagepagefraction{1}
\def\textpagefraction{.001}
\shorttitle{Financial Ticket Faster Recognition}
\shortauthors{F Tian et~al.}

\title [mode = title]{Research on Fast Text Recognition Method for Financial Ticket Image}          

\author[1]{Fukang Tian}[style=chinese]
\author[1]{Haiyu Wu}[style=chinese]
\author[1]{Bo Xu}[style=chinese, orcid=0000-0002-7304-857X]
\cormark[1]

\address[1]{Xi'an Network Computing Data Technology Co., Ltd., Xi'an 710049, China}

\cortext[cor1]{Corresponding author}

\begin{abstract}
Currently, deep learning methods have been widely applied in and thus promoted the development of different fields. In the financial accounting field, the rapid increase in the number of financial tickets dramatically increases labor costs; hence, using a deep learning method to relieve the pressure on accounting is necessary. At present, a few works have applied deep learning methods to financial ticket recognition. However, first, their approaches only cover a few types of tickets. In addition, the precision and speed of their recognition models cannot meet the requirements of practical financial accounting systems. Moreover, none of the methods provides a detailed analysis of both the types and content of tickets. Therefore, this paper first analyzes the different features of 482 kinds of financial tickets, divides all kinds of financial tickets into three categories and proposes different recognition patterns for each category. These recognition patterns can meet almost all types of financial ticket recognition needs. Second, regarding the fixed format types of financial tickets (accounting for 68.27\% of the total types of tickets), we propose a simple yet efficient network named the Financial Ticket Faster Detection network (FTFDNet) based on a Faster RCNN. Furthermore, according to the characteristics of the financial ticket text, in order to obtain higher recognition accuracy, the loss function, Region Proposal Network (RPN), and Non-Maximum Suppression (NMS) are improved to make FTFDNet focus more on text. Finally, we perform a comparison with the best ticket recognition model from the ICDAR2019 invoice competition. The experimental results illustrate that FTFDNet increases the processing speed by 50\% while maintaining similar precision.
\end{abstract}


\maketitle

\section{Introduction}
Rapidly developed computer hardware allows deep learning methods to be widely implemented and studied. Currently, deep learning methods have gradually released people from heavy and repetitive work. Financial accounting is an important field that mainly relies on manual work. Traditional accounting is usually performed in the following steps. First, accountants need to sort different types of financial tickets, such as Value-Added Tax (VAT) invoices, bank tickets, toll tickets, taxi tickets, train tickets, etc. Then, they enter the basic information of these tickets into financial software to produce accounting vouchers for the corresponding category. Subsequently, each financial ticket is sequentially attached to the accounting voucher for the corresponding category. Finally, the accountant must double check if there are any missing tickets and if the sequence of tickets is correct. It is obvious that the traditional approach lacks automation and has a considerable amount of redundant work. Hence, considering the massive recognition workloads, time consumption, and labor efforts of accountants, a large number and variety of financial tickets leads to high labor costs and low work efficiency. Therefore, in order to make financial accounting more accurate, more efficient, and highly automated, optical character recognition (OCR) technology has been somewhat implemented in industry.

Processing and analyzing ticket images to automatically obtain ticket text information can effectively reduce the manual labor required for financial ticket audit and entry work and improve the efficiency of financial accounting work. At present, deep learning enables optical character recognition to overcome most of the noise interference and achieve unprecedented accuracy. It also enables ticket recognition to meet the actual requirements of high accuracy in financial work under certain conditions. However, the following problems still exist: 1) The ticket recognition algorithm based on deep learning has a large amount of computations and is time-consuming. In addition, the audit business of financial tickets has the characteristics of a large number of tickets and high concurrency, which makes the running speed of the recognition algorithm the bottleneck of its practical application. 2) The financial industry is less tolerant of errors; and in many cases, subtle differences lead to very different results. Most ticket recognition algorithms have three stages or more, and the errors in each stage have a cumulative effect. Therefore, improving the ticket recognition accuracy is still the core problem. 3) At present, most of the common ticket recognition algorithms use one or a few types of financial tickets for research and lack systematic research on all financial ticket data. Hence, there is still a considerable amount of room to improve the speed and accuracy of the algorithm according to the overall characteristics of the ticket data.

In order to solve the above problems, this paper proposes a fast recognition algorithm for full sample financial tickets based on 482 types of 1.1 million financial tickets in different regions of China. Moreover, it is worth mentioning that a “ticket” can be an invoice, receipt, ticket, bill, etc. in this paper. Our contributions are as follows:
\begin{itemize}
    \item Based on the sufficient database, we summarized 482 types of financial tickets into two types. For each type, we design different patterns according to the characteristics of the different types of financial tickets so as to greatly improve the accuracy and speed of financial ticket detection and recognition and achieve efficient text detection and recognition.
    \item We propose a simple yet efficient network named the Financial Ticket Faster Detection network (FTFDNet), which is used to recognize fixed form types of financial tickets (accounting for 68.27\% of the total types of tickets) to achieve a more efficient and accurate recognition effect.
    \item We improved the loss function, RPN, and NMS of FTFDNet in order to obtain more accurate results and make FTFDNet focus more on text. The experimental results prove the effectiveness of these improvements.
\end{itemize}
\section{Related work}

\subsection{Object detection}

Current object detection algorithms can be divided into one-stage structures and two-stage structures. Two-stage networks, such as the Faster RCNN\cite{ren2015faster}, have high detection accuracy, but they do not perform well in real-time detection and video detection, which require high detection speed. Therefore, one-stage networks such as YOLOs\cite{redmon2016you,redmon2017yolo9000,redmon2018yolov3,bochkovskiy2020yolov4} have been proposed. They have made great improvements in detection speed while maintaining high accuracy. However, in relatively special fields, such as text detection, a general detection network cannot adequately conduct detection and recognition because of the special scene of text detection and recognition and the requirements for precision being more stringent.

\subsection{Text detection and recognition}
In view of the particularity of text, many networks have been proposed. The CTPN \cite{tian2016detecting} focuses on horizontal direction text detection; EAST \cite{zhou2017east} improves the detection efficiency; and RARE \cite{shi2016robust}, Mask TextSpotter \cite{lyu2018mask}, ABCNet \cite{liu2020abcnet} and others are designed to detect arbitrarily shaped text such as bending. In addition, they improved the text detection method in speed and accuracy. Because different combinations of words represent different meanings, it is necessary to recognize the text content. CRNN \cite{shi2016end}, Seq2seq \cite{sutskever2014sequence}, and Transformer \cite{vaswani2017attention}, as representative text recognition methods, have been widely used in various fields. There are also some studies on the alignment of detected text.

\subsection{Ticket recognition}
Currently, text recognition has been applied to various fields. Financial tickets, as transaction vouchers, are a very important application spot. The diversity of the categories and the complexity of the contents of financial tickets force companies to use text detection and recognition technology to reduce labor costs. However, due to the large number of tickets closely related to funds, the requirements for the recognition accuracy and speed are high. \cite{sun2019template,zhang2019research,cesarini2003analysis,palm2017cloudscan, blanchard2019automatic, yi2019dual,klein2004results, kieri2012context} used an RNN, LSTM, an AA-RNN to recognize medical tickets and VAT invoices. However, as mentioned above, due to the diversity of types and complexity of content, these models cannot include all types of tickets. Therefore, we propose three different modules to handle most ticket detection and recognition tasks according to the types and contents of tickets.

\section{Tickets and recognition patterns}
Financial ticket recognition has unique business requirements. In the process of accounting entry, different information needs to be recognized from different tickets. Therefore, the algorithm logic process can be adjusted according to the actual business needs so as to improve the operating speed of the algorithm.

\subsection{Analysis of recognition time}
The financial work application information in the ticket can be expressed as:
\begin{equation}
    I = \{K_{1} : V_{1}, K_{2} : V_{2}, ..., K_{n} : V_{n}\}
\end{equation}
where $K_{n}$ is the keyword category of the $n_{th}$ region-of-interest (ROI) area in the invoice image; and $V_{n}$ is the string content corresponding to the region, which is composed of all character position information and content information in the ROI area. $V_{n}$ is expressed as:

\begin{equation}
    V_{n} = \sum_{roi}P_{char} + \sum_{roi}C_{char}
\end{equation}

Therefore, the fundamental task of financial ticket recognition is to extract key information such as $K$, $P$ and $C$ from a ticket image. In addition, the time consumption of single sample recognition can be expressed as follows:
\begin{equation}
    T = \alpha \times (w + h) + \beta \times A_{text} + \gamma \times A_{information} + t
\end{equation}

where $(w+h)$ is the resolution of the sample image; $A_{text}$ is the sum of the text area of the image; $A_{information}$ is the sum of the business information areas in the image; and $t$ is the wear constant, which represents the lost time from the reading, transmission and structure of the system and is determined by the operating environment of the recognition system and the efficiency of the structured algorithm. $\alpha$, $\beta$, and $\gamma$ are the coefficients of each respectively factor. The formula shows that the recognition time of a single sample is proportional to the sample image size, text area and information area.

\subsection{Connections between types and patterns}

\begin{table}
\caption{According to the differences of the entry information, tickets could be sorted into two categories.}
\begin{center}
    \begin{tabular}{ccccc}
    \hline
    \multirow{2}*{Ticket type} & \multicolumn{2}{c}{Text form} & \multicolumn{2}{c}{Text vocabulary}\\
    \cline{2-5}
    & fixed & non-fixed & fixed & non-fixed\\
    \hline
    I-A & \textbf{\checkmark} & & &\textbf{\checkmark}\\
    \hline
    I-B & \textbf{\checkmark} & & \textbf{\checkmark}  &\\
    \hline
    II & & \textbf{\checkmark} & \textbf{\checkmark} &\\
    \hline
    \end{tabular}
\end{center}
\label{table1}
\end{table}

\begin{figure}
\begin{center}
\centerline{\includegraphics[width=1\columnwidth]{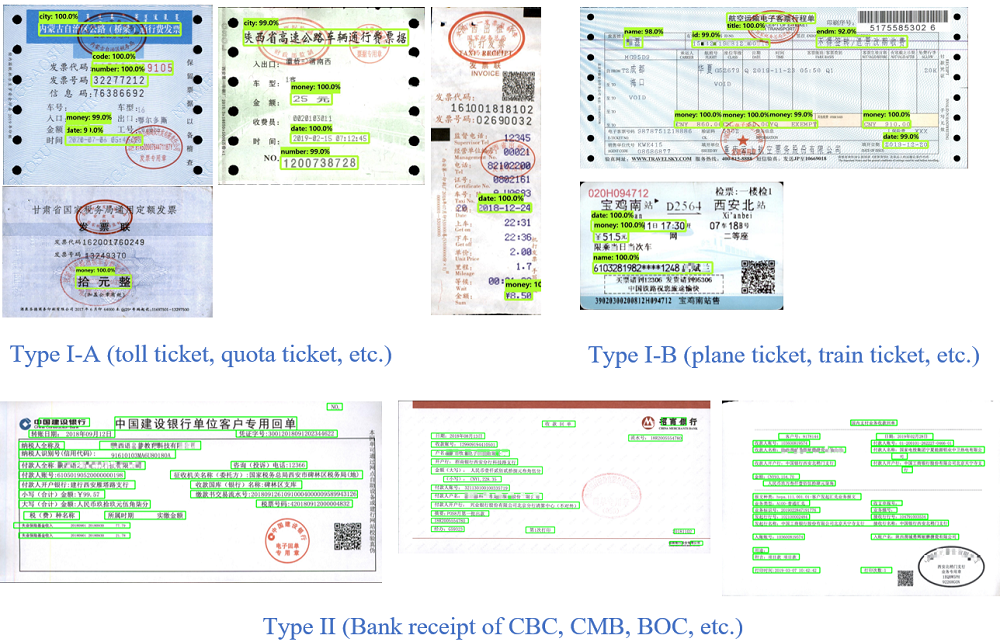}}
\caption{Type I-A is the ticket with fixed simple forms. Type I-B is the ticket with fixed complex forms. Type II is the ticket with non-fixed forms. It is obvious that type II include more diverse contents.}
\label{fig1}
\end{center}
\end{figure}

As shown in Table\ref{table1}, according to the differences in the information required when an entry is recorded, financial tickets can be divided into two types. \textbf{I}. Fixed form type. The recognition content of this type of ticket is a fixed form target. Specifically, it can be divided into two subtypes based on the vocabulary. A) Simple vocabulary types, such as toll invoices, taxi invoices, quota invoices, value-added tax invoices, etc. The target form usually includes a date, an amount, an invoice code, an invoice number, a verification code, etc.; and the recognition content is specific Chinese characters, English letters or numbers. B) Complex vocabulary type. The recognition content of this type of ticket is also a fixed form field, but because the target contains a name field, it involves a complex vocabulary of Chinese characters. Examples include train tickets, plane itinerary tickets, etc. The target area of this type of ticket contains the person’s name area, so the characters to be recognized could be one of more than 4000 common Chinese characters, which makes it difficult to recognize the target information directly by using FTFDNet. Therefore, it is necessary to add a Chinese character recognition model on the basis of FTFDNet to improve the overall accuracy. \textbf{II}. Types of non-fixed forms. The recognition contents of some tickets are non-fixed form fields, such as bank receipts and voucher tickets. The recognition target forms vary, and the recognition contents include Chinese characters, English letters, punctuation marks, special symbols and other characters. Through the statistical analysis of 716872 tickets produced by 276 companies in China in 2019, type I tickets accounted for 68.27\% and type II tickets accounted for 31.73\% of the total. Fig.\ref{fig1} shows some examples of Type I and Type II tickets.

\begin{figure}
\begin{center}
\centerline{\includegraphics[width=1\columnwidth]{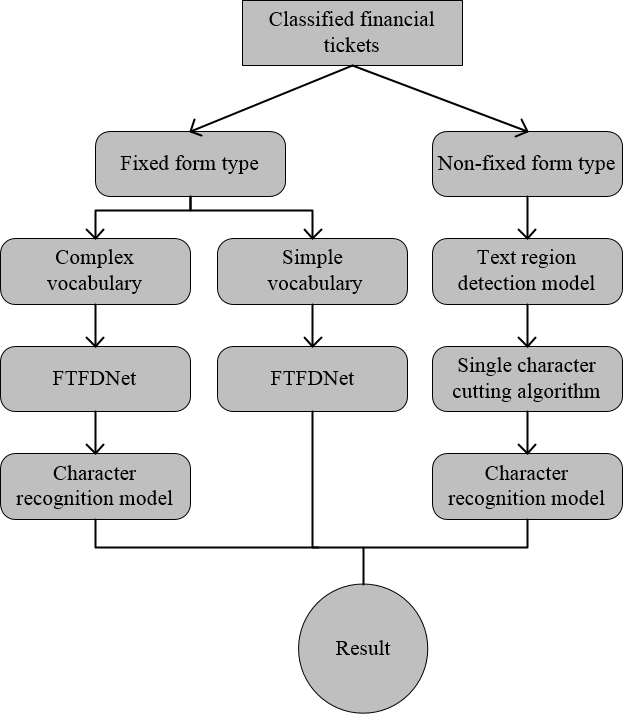}}
\caption{The flowchart for the financial ticket recognition pattern, where the most left line is recognition pattern I, the middle line is recognition pattern II, and the right line is recognition pattern III.}
\label{fig2}
\end{center}
\end{figure}

In view of the above two types of tickets, we propose recognition patterns according to their characteristics. The flowchart of these three recognition patterns is shown in Fig.\ref{fig2}.

\textbf{Recognition patterns I and II}: For Type I tickets, containing Type I-A and Type I-B, the financial accounting contents are relatively fixed; thus, FTFDNet is used to directly detect and recognize the needed information. Nevertheless, because some Chinese characters are included in type I-B tickets, if the same pattern is used to recognize a large vocabulary of characters, the accuracy cannot match the requirement. Hence, a character-level recognition model is added to recognize Chinese characters.

\textbf{Recognition patterns III}: For Type II tickets, financial accounting has large differences, and full surface detection is needed to extract the required information. However, if the end-to-end text recognition model, such as a CRNN, is directly used to recognize the detection result, the amount of data annotation for training is too large, which will require high labor costs to make this dataset. Therefore, we design a pattern for it, which cuts the detection results at the character level; and finally use the character-level recognition model to recognize the content.

For patterns I and II, tickets occupy 68.27\% of the whole dataset. Hence, reducing the ticket recognition time would significantly improve the overall performance of the method in daily business. Therefore, we propose a simple yet efficient network named the FTFDNet based on the Faster RCNN. First, the Faster RCNN is used to detect the ROI region to be recognized and the information $K$ of each region in the whole ticket. Then, single character target detection is applied to the cut ROI region image to determine the information such as $P$ and $c$ in each ROI region, and the keyword information and superposition detection results are integrated to form the final word recognition result. In addition, according to the characteristics of some ticket recognition targets with relatively fixed shapes and small vocabularies, key information such as $K$, $P$ and $c$ is determined in the superimposed detection process, which eliminates the character recognition model and result construction, greatly reduces $\alpha$ and $c$, and improves the overall ticket recognition speed.
\section{Financial ticket Faster detection network FTFDNet}
\subsection{Network structure}

\begin{figure*}
\begin{center}
\centerline{\includegraphics[width=2\columnwidth, height=10cm]{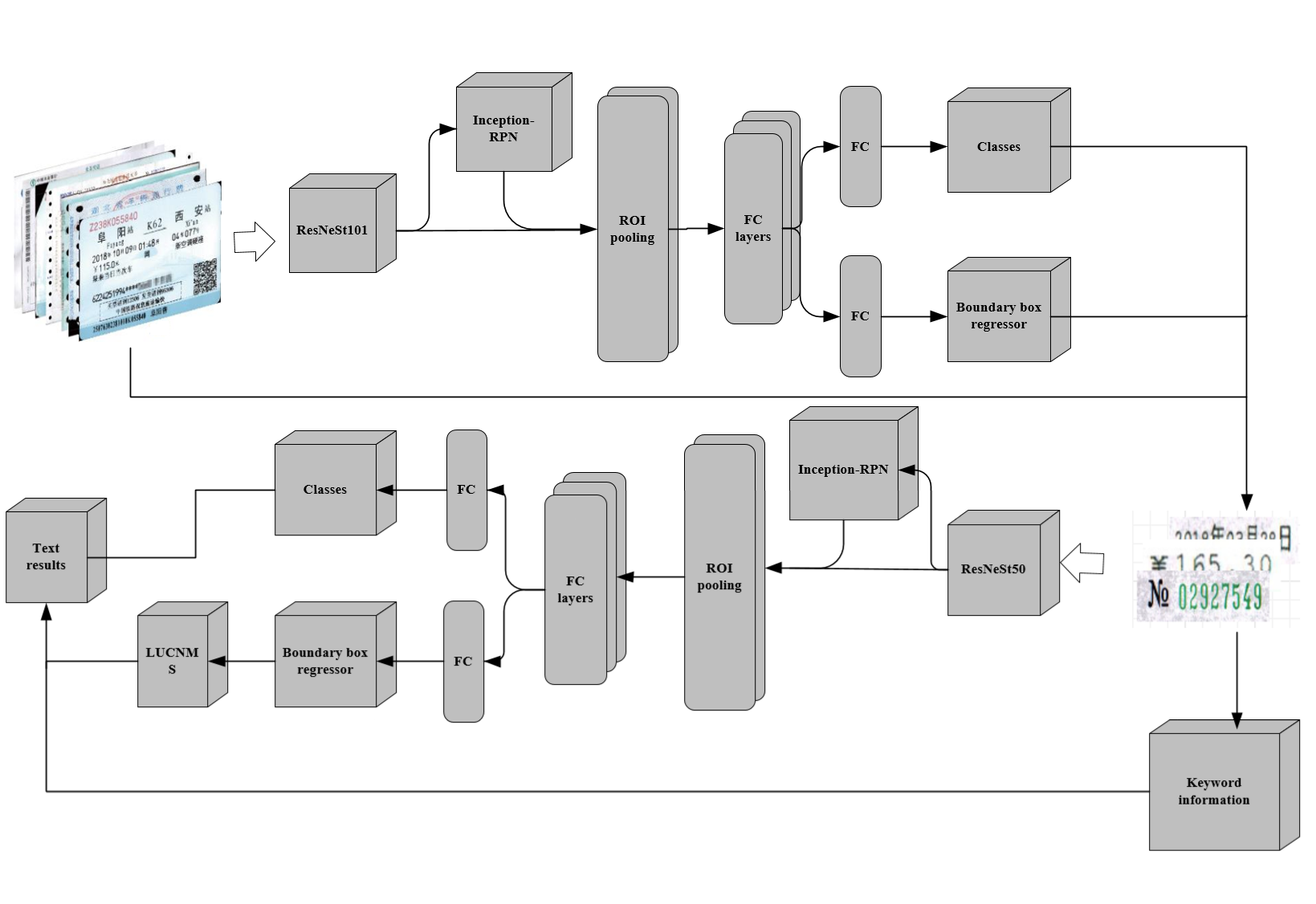}}
\caption{It shows the structure of FTFDNet, which contains the text region detection part and character information extraction part.}
\label{fig6}
\end{center}
\end{figure*}

According to the characteristics of financial ticket data, the financial ticket detection network is a special network that can quickly extract the information of type I and II financial tickets in fewer steps based on the improved object detection model. Its structure is shown in Fig.\ref{fig6}.

\begin{table*}[!hbt]
\caption{THE RESULTS OF COMPARING THE CORNER+CRNN AND FTFDNET ON THE TAXI TICKET, TRAIN TICKET, VAT TICKET, AND QUOTA TICKET DATASET.}
\begin{center}
    \begin{tabular}{cccccccccc}
    \hline
    \multirow{2}*{Type} & \multirow{2}*{Resolution} & \multicolumn{2}{c}{Detection AP50} & \multicolumn{2}{c}{D-time(FPS)} & \multicolumn{2}{c}{Recognition AP50} & \multicolumn{2}{c}{R-time(FPS)}\\ \cline{3-10}
    & & FTFDNet & Corner & FTFDNet & Corner & FTFDNet & CRNN & FTFDNet & CRNN \\ \hline
    VAT tickets & $1024\times2048$ & 99.8 & $97.6$ & \textbf{22.22} & $0.3$ & $96.2$ & $96.5$ & $38.46$ & \textbf{47.63} \\
    Taxi tickets & $600\times1024$ & 97.5 & $98.9$ & \textbf{34.48} & $1.64$ & $99$ & $97.4$ & $41.67$ & \textbf{50} \\
    Quota tickets & $600\times1024$ & 99.6 & $98.9$ & \textbf{35.71} & $1.27$ & $99.5$ & $97.6$ & \textbf{45.46} & $45.43$ \\
    Train tickets & $520\times1500$ & 99.7 & $99.5$ & \textbf{37.04} & $2.12$ & $99.3$ & $96.1$ & $40$ & \textbf{48.21} \\ \hline
    \end{tabular}
\end{center}
\label{table4}
\end{table*}

The first step is text area detection. Image features are extracted by using ResNeSt101 \cite{zhang2020resnest} as the backbone, which has a strong ability to extract the features; and the text region position is detected by Inception-RPN \cite{zhong2016deeptext}, which is designed to fit the text characteristics. In addition, the structures of ResNeSt and Inception-RPN are introduced in \ref{backbone} and \ref{RPN}, respectively. The data features of the ROI region are standardized by multilevel region-of-interest pooling, and then financial keyword classification and location regression are conducted. In this step, the financial business keyword location $K$ and ROI image will be given. The second step is character information extraction. ResNeSt50 is used as the backbone to extract ROI image features, and the RPN network is used to determine the character position information. ROI pooling is used to standardize the data features of a single character area. Then, character classification and position regression are conducted, that is, $P$, $C$, etc. in the ROI image. It is worth mentioning that in order to improve the detection accuracy of FTFDNet, the CIoU loss and local-based unique character NMS are implemented. The details are in \ref{loss} and \ref{NMS}, respectively. Finally, financial information $I$ in the whole ticket is obtained by combining the information $K$ obtained in the first step.

\subsection{Backbone ResNeSt}\label{backbone}

\begin{figure}
\begin{center}
\centerline{\includegraphics[width=1\columnwidth]{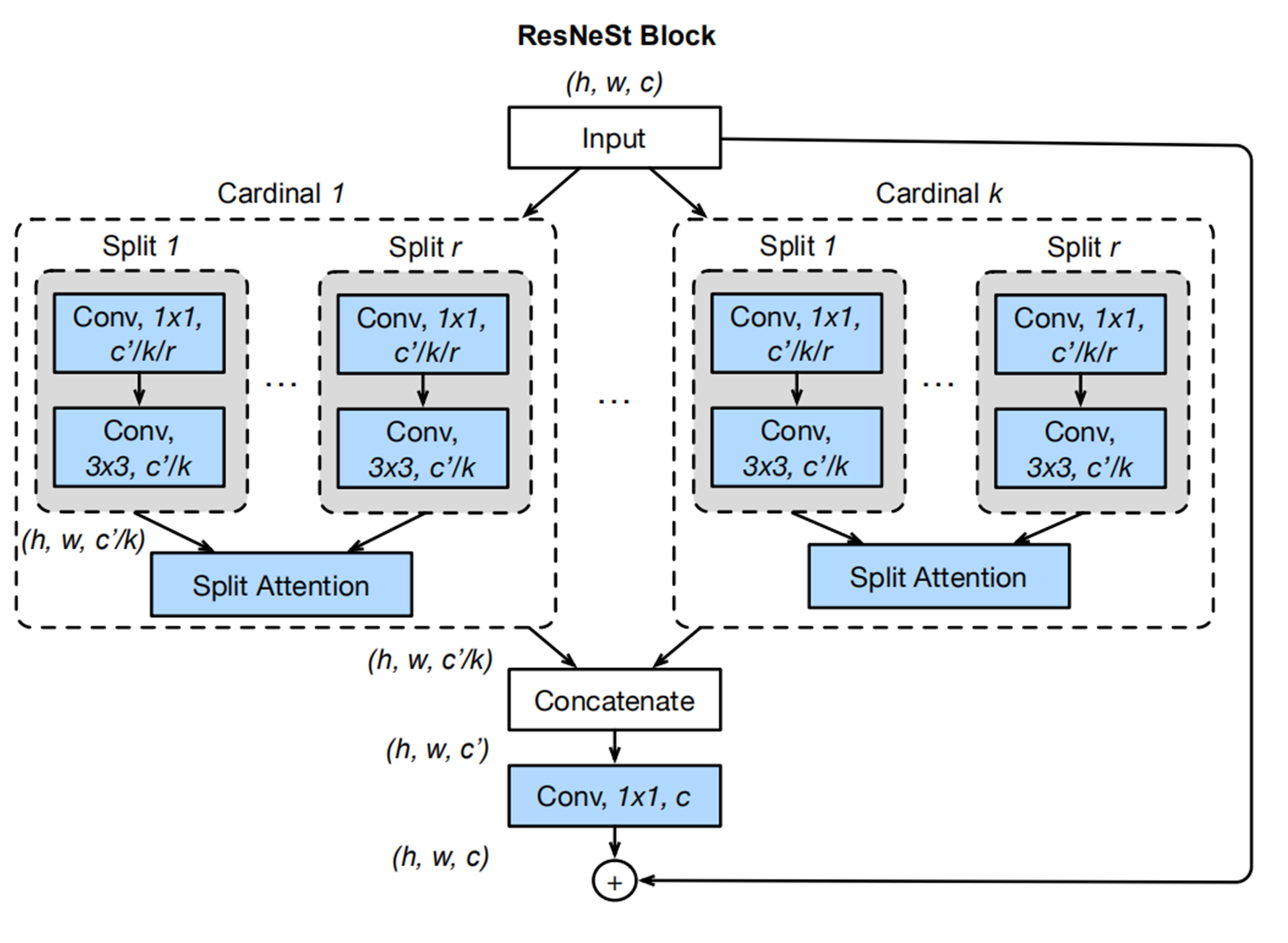}}
\caption{The main structure of the ResNeSt block.}
\label{fig3}
\end{center}
\end{figure}

Due to the strict requirements for the ticket recognition accuracy, the backbone is required to have very outstanding performance to provide a better backup for the subsequent detection and recognition heads. Therefore, we chose the ResNeSt backbone recently proposed by \cite{zhang2020resnest}. The framework is shown in Fig.\ref{fig3}. First, the main structure is consistent with ResNeXt \cite{xie2017aggregated} in order to widen the width of the network and conduct multigroup feature fusion. This structure can greatly improve the feature extraction ability in each layer. In the feature extraction stage of each group, the network adopts the idea of GoogLeNet GoogLeNet\cite{szegedy2015going} and uses multiscale convolution. The combination of different convolution sizes can extract more image features. Finally, all the extracted features are determined by the channel attention, which makes the important features extracted more comprehensive and accurate. Therefore, it is more powerful for the feature extraction of Chinese text with complex structures. Moreover, its model parameters are similar to ResNet \cite{he2016deep} with the same depth, and the amount of content to be identified in the recognition phase of FTFDNet is less, which could improve the operating speed. Therefore, in order to ensure a higher recognition speed, we adopt ResNeSt50 in the second stage of FTFDNet. The comparative test is shown in Table\ref{table2}. 

\subsection{Area detection network Inception-RPN}\label{RPN}

\begin{figure}[htbp]
\begin{center}
\centerline{\includegraphics[width=1\columnwidth]{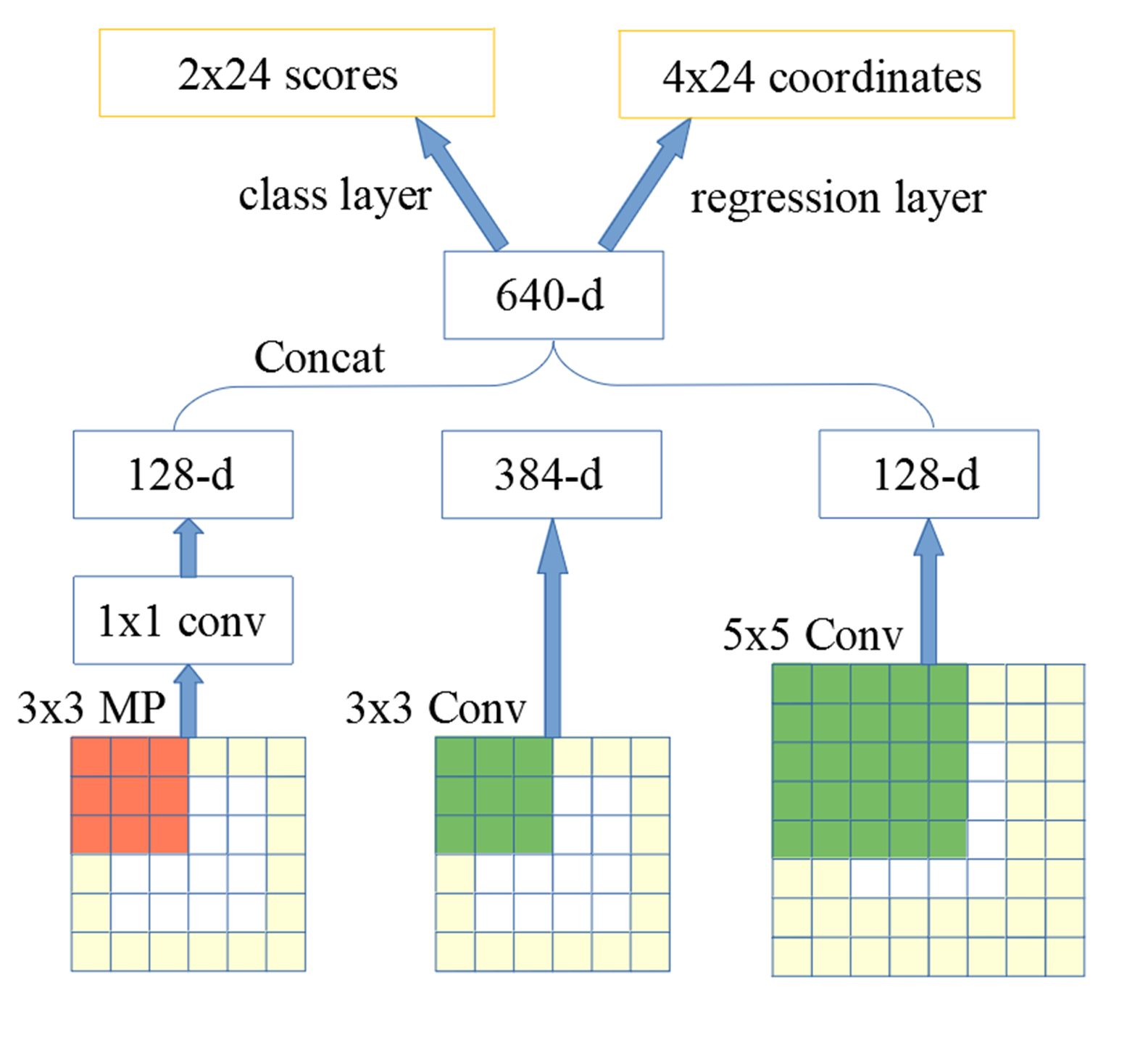}}
\caption{The main structure of Inception-RPN.}
\label{fig4}
\end{center}
\end{figure}

The shape of the text area is mostly a slender area with a width greater than its length, which is a significant feature different from general object detection. In order to accurately detect its position information, we select the Inception-RPN network proposed by \cite{zhong2016deeptext} as the deep text network, and its structure is shown in Fig.\ref{fig4}. According to the shape features of the text area, the network extends nine anchors with three levels and three scales on each pixel to four levels (32, 48, 64, and 80) and 24 anchors in 6 scales (0.2, 0.5, 0.8, 1.0, 1.2, and 1.5) to make it more suitable for text region detection. In addition, the network introduces the inception structure of GoogLeNet, uses a $3\times3$ convolution kernel, a $5\times5$ convolution kernel and $3\times3$ max pooling to extract the local features, and then forms a 640-d feature vector for text region classification and position regression. Therefore, the Inception-RPN has the following advantages: 1) The multiscale convolution features are conducive to foreground and background classification. 2) Convolution and pooling can effectively extract the regional features of text information.

\subsection{Loss function}\label{loss}
In object detection, the loss function generally used in the regression of the predicted bounding box is the smooth L1 loss \cite{ren2015faster}.
 
\begin{equation}
\begin{aligned}
& t_{x} = (x - x_{a}) / w_{a}, t_{y} = (y - y_{a}) / h_{a},\\
& t_{w} = \log{w/w_{a}}, t_{h} = \log{h/h_{a}},\\
& t_{x}^{*} = (x^{*} - x_{a}) / w_{a},  t_{y}^{*} = (y^{*} - y_{a}) / h_{a},\\
& t_{w}^{*} = \log{w^{*}/w_{a}}, t_{h}^{*} = \log{h^{*}/h_{a}},\\
\end{aligned}
\end{equation}
Where $x$, $y$, $w$, and $h$ are the central point coordinates, width, and height of the box. As for $x$, $x_{a}$, and $x^{*}$ are for the predicted box, anchor box, and ground truth box respectively, similarly for $y$, $w$, and $h$. However, it can be seen from the above formula that the four vertex coordinates of the predicted bounding box are used in the regression of the prediction box, and the loss of the four points is independently calculated. This operation ignores the relationship between the four vertices, which will cause the loss of multiple predicted bounding box to be similar, but the difference of IoU will be very large, which has a great impact on the detection and recognition results. IoU can directly reflect the fit degree of predicted bounding box and ground truth, so IoU loss, GIoU loss\cite{rezatofighi2019generalized}, PIoU loss\cite{chen2020piou}, and CIoU loss\cite{zheng2020distance} have better performance in detection. Among them, CIoU loss 

where $x$, $y$, $w$, and $h$ are the central point coordinates, width, and height of the box, respectively. $x$, $x_{a}$, and $x^{*}$ are the predicted box, anchor box, and ground truth box, respectively, similar to $y$, $w$, and $h$. However, it can be seen from the above formula that the four vertex coordinates of the predicted bounding box are used in the regression of the prediction box, and the loss of the four points is independently calculated. This operation ignores the relationship between the four vertices, which will cause the loss of multiple predicted bounding boxes to be similar, but the difference in the IoU will be very large, which has a great impact on the detection and recognition results. The IoU can directly reflect the degree of fit of the predicted bounding box and ground truth so that the IoU loss, GIoU loss \cite{rezatofighi2019generalized}, PIoU loss \cite{chen2020piou}, and CIoU loss \cite{zheng2020distance} have better detection performance. Among them, the CIoU loss is used here.

\begin{equation}
\begin{aligned}
& L_{CIoU} = 1 - IoU + \frac{p^{2}(b,b^{gt})}{c^{2}} + av\\
& v = \frac{4}{\pi}(arctan\frac{w^{gt}}{h^{gt}}- arctan\frac{w}{h})\\
\end{aligned}
\label{equation5}
\end{equation}

Formula\ref{equation5} calculates the aspect ratio, IoU, and central points, which could make the prediction box and ground truth more consistent. Therefore, in order to ensure the high accuracy of ticket recognition results, we choose the CIoU loss. The results of the comparative test are shown in Table\ref{table3}.

\subsection{Location based Unique Character NMS}\label{NMS}

\begin{figure}
\begin{center}
\centerline{\includegraphics[width=1\columnwidth]{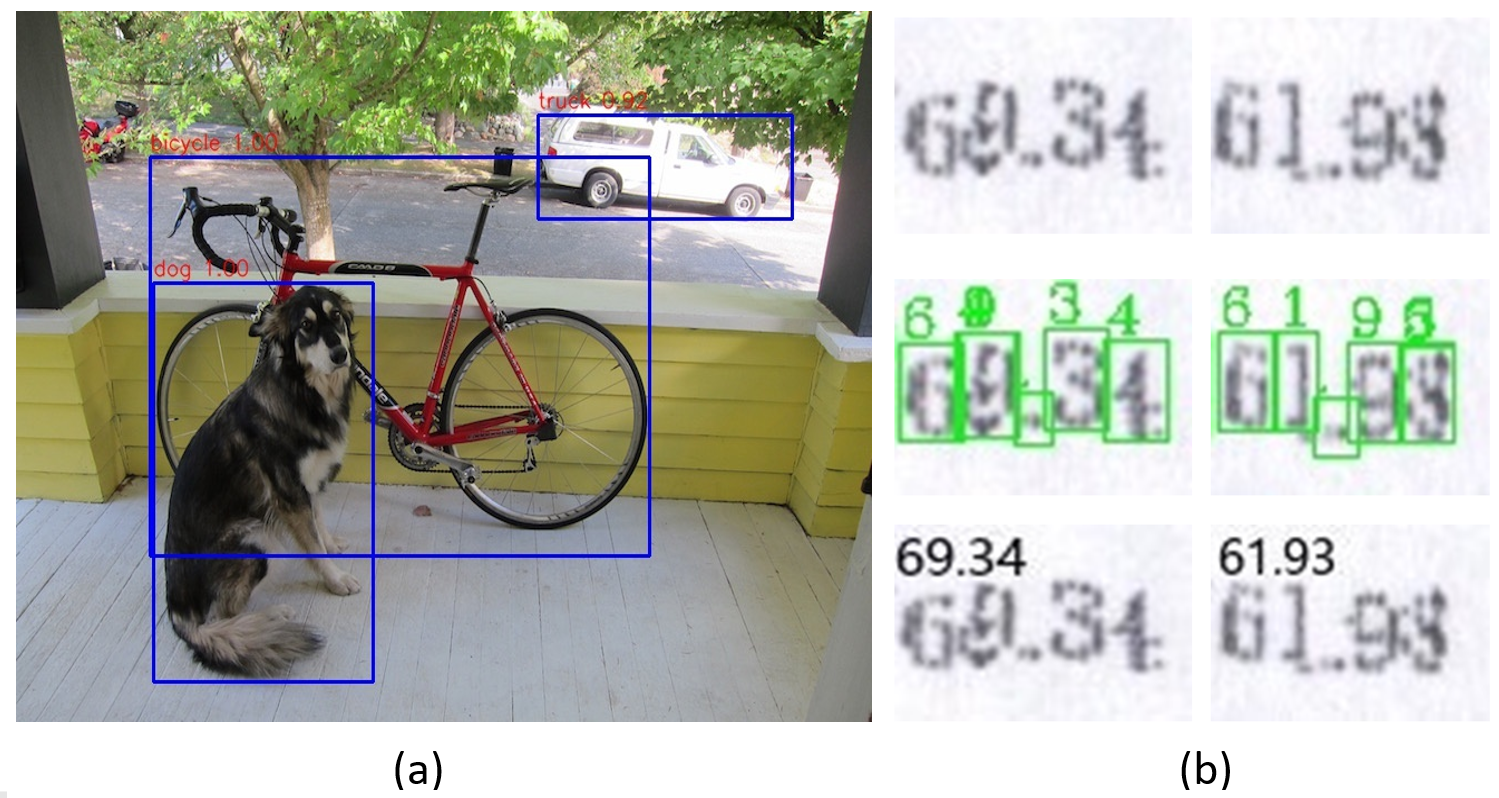}}
\caption{It gives an example showing the difference between general object detection and text detection.}
\label{fig5}
\end{center}
\end{figure}

The Faster RCNN is used for the object detection of general data. A common data set has the situation that different kinds of objects overlap, as shown in Fig.\ref{fig5}a. In terms of the detection and recognition of line text, words are not allowed to overlap. Therefore, using the general NMS algorithm may cause multiple text detection results in one position, as shown in Fig.\ref{fig5}b, which will have a great impact on the ticket detection and recognition results. Therefore, in view of this phenomenon, we improve the traditional NMS algorithm so that the output results in a similar position will only appear as one result; thus, LUCNMS can greatly improve our detection and recognition accuracy.

\section{Experiments}
\subsection{Datasets}
In this paper, we use 1226 taxi tickets, 436 train tickets, 784 quota tickets and 4484 VAT tickets to train Corner \cite{lyu2018multi} + CRNN and FTFDNet. In addition, 1600 tickets, 400 for each category, are used for testing. For the unilateral improved comparative test, including backbones and loss functions, 4184 VAT tickets and 16146 ROI regional data are used to train the detection and recognition models, respectively. A total of 700 VAT tickets and 1300 ROI regional data points were used for testing.

\subsection{Implementation}
We use the MMDetection and Detectron2 object detection frameworks to test the different structures of the Faster RCNN in the comparative experiment based on backbones and loss functions. In order to ensure the effectiveness of the experiment, we choose the default hyperparameter settings of the frameworks, do not use any tricks, and then select the optimal result. In order to accelerate the convergence of the model, we add the COCO2017 pretrained model. As for the comparison of the best model, FTFDNet is the final optimized model, which integrates ResNeSt101, the CIoU loss, the Inception-RPN, and the location-based unique character NMS. We use a single Tesla P40 24 GB GPU to perform the training and testing work.

\subsection{Backbones}
We compare the performance of ResNet101, ResNeXt101, Res2Net101 [31] and ResNeSt101 as the backbones for text detection and recognition on the VAT ticket dataset. The results are shown in Table\ref{table2}. ResNeSt101 yields 99.51\% keyword box detection and 94.1\% on char box detection. Compared with the other three backbones, ResNeSt101 could provide the highest detection and recognition accuracy, which is effective for reducing the negative cases when the number of tickets is large.

\begin{table}
\caption{The comparison results, based on AP50, of FTFDNet when ResNet and its variants are respectively used as the backbone.}
\begin{center}
    \begin{tabular}{cccc}
    \hline
    Network & Backbones & Keyword box & char box\\
    \hline
    \multirow{4}*{Faster RCNN} & ResNet101 & 99.49 & 93.8\\ \cline{2-4}
    & ResNeXt101 & 99.47 & 92.8\\ \cline{2-4}
    & Res2Net101 & 99.30 & 93.0\\ \cline{2-4}
    & ResNeSt101 & \textbf{99.51} & \textbf{94.1}\\ \hline
    \end{tabular}
\end{center}
\label{table2}
\end{table}

\subsection{Loss function}
In this part, we compare the loss functions of the IOU series. Since the text content of the ticket is horizontal line text, we  did not test the PIoU loss function for the target with a rotation angle. From Table\ref{table3}, the CIoU loss has the best bounding box regression results on the char box, which also proves the analyses in Section 3.

\begin{table}
\caption{Comparison results of FTFDNet when the bounding box regression loss function is Smooth L1, IoU, GIoU, and CIoU, respectively.}
\begin{center}
    \begin{tabular}{cccc}
    \hline
    Network & Loss function & AP50 & Recall\\
    \hline
    \multirow{4}*{Faster RCNN+ResNet101} & Smooth L1 & 93.8 & 78.9\\ \cline{2-4}
    & IoU & 94.2 & 80.1\\ \cline{2-4}
    & GIoU & 94.2 & 79.5\\ \cline{2-4}
    & CIoU & \textbf{94.5} & \textbf{81.2}\\ \hline
    \end{tabular}
\end{center}
\label{table3}
\end{table}

\subsection{Comparison with Corner+CRNN}
To verify the performance of FTFDNet, we choose Corner+CRNN, which is the champion model of the ICDAR2019 invoice end-to-end detection and recognition competition, to perform a comparative test on the VAT dataset. Furthermore, in this experiment, the training tricks are the same, and the parameter settings are set as default. The metrics are the recognition speed and recognition accuracy. From Table \ref{table4}, our FTFDNet maintains a high recognition speed with comparable precision, which also verifies the original intention of our network design.
\subsection{Comparison with related methods}
All the research and studies are mainly aimed at a few kinds of tickets. For instance, \cite{liu2020end} uses the SSD \cite{liu2016ssd} and the CNN-GRU \cite{chung2014empirical} to detect and recognize taxi receipts, and the accuracy of their models reached 94.36\%. More comparison results are shown in table\ref{table5}. The table shows that our method is a great leap forward in the field of ticket recognition.

\begin{table}
\caption{It shows the supported types of tickets for each method and the recognition accuracy}
\begin{center}
    \begin{tabular}{cccc}
    \hline
    Methods & \# of types & Ticket types & Accuracy\\
    \hline
    Liu\cite{liu2020end} & 1 & taxi & 94.36\% \\
    \hline
    \multirow{2}{*}{Yang\cite{yang2019deep}} & \multirow{2}{*}{3} & VAT, train,  & \multirow{2}{*}{97.1}\% \\
    &  & ordinary machine &\\
    \hline
    Zhang\cite{zhang2019research} & 1 &VAT & 96.21\% \\
    \hline
    \multirow{2}{*}{Ours} & \multirow{2}{*}{\textbf{194}} & VAT, taxi, train, & \multirow{2}{*}{\textbf{97.4\%}} \\
    &  &  bank receipt... &\\
    \hline
    \end{tabular}
\end{center}
\label{table5}
\end{table}

\section{Conclusion}
Based on the analyses of financial tickets, we divide them into two types. Then, the three different patterns of text detection and recognition are designed for type I-A, type I-B and type II. Currently, these three patterns can handle all kinds of tickets. We propose a simple yet efficient network model, FTFDNet, to detect and recognize the fixed form types of  tickets. To make this model more suitable for fitting the text characteristics, we use the CIoU loss, LUCNMS, Inception-RPN, and ResNeSt to enhance the performance. Finally, compared with the best detection and recognition model in the ICDAR2019 invoice competition, FTFDNet can maintain comparable precision, which also verifies the original intention of our network design. Finally, compared with other methods, our methodology is a huge leap forward in ticket recognition.

\bibliographystyle{cas-model2-names}

\bibliography{cas-refs}


\end{sloppypar}
\end{document}